\documentclass{bmvc2k}

\title{Learning Object-level Point Augmentor for Semi-supervised 3D Object Detection}

\addauthor{Cheng-Ju Ho*}{ace52751208@gmail.com}{1}
\addauthor{Chen-Hsuan Tai*}{derek.0417t@gmail.com}{1}
\addauthor{Yi-Hsuan Tsai$^\ddagger$}{wasidennis@gmail.com}{2}
\addauthor{Yen-Yu Lin}{lin@cs.nycu.edu.tw}{1}
\addauthor{Ming-Hsuan Yang}{mhyang@ucmerced.edu}{3}

\addinstitution{
 National Yang Ming Chiao Tung \\
 University, Taiwan
}
\addinstitution{
Phiar Technologies, \\
United States
}
\addinstitution{
University of California at Merced, \\
United States
}

\runninghead{C.-J. Ho et al.}{Learning Object-level Point Augmentor}

\def\ie{\emph{i.e}\bmvaOneDot}

\newcommand{\ourname}{OPA}

\newcommand{\myfootnote}[2]{{%
  \let\thempfn\relax
  \footnotetext[0]{$^#1$\emph{#2}}
}}

\usepackage{multirow}
\usepackage{tabularx}
\usepackage{booktabs}
\usepackage{amsfonts}
\usepackage{graphicx}
\usepackage{comment}
\usepackage{amsmath,amssymb} 
\usepackage{color}
\usepackage{graphicx}
\usepackage{enumerate}

\begin{document}

\maketitle

\begin{abstract}
Semi-supervised object detection is important for 3D scene understanding because obtaining large-scale 3D bounding box annotations on point clouds is time-consuming and labor-intensive.
Existing semi-supervised methods usually employ teacher-student knowledge distillation together with an augmentation strategy to leverage unlabeled point clouds.
However, these methods adopt global augmentation with scene-level transformations and hence are sub-optimal for instance-level object detection.
In this work, we propose an object-level point augmentor ({\ourname}) that performs local transformations for semi-supervised 3D object detection.
In this way, the resultant augmentor is derived to emphasize object instances rather than irrelevant backgrounds, making the augmented data more useful for object detector training. 
Extensive experiments on the ScanNet and SUN RGB-D datasets show that the proposed {\ourname} performs favorably against the state-of-the-art methods under various experimental settings. 
%
The source code will be available at \url{https://github.com/nomiaro/OPA}.
%

\end{abstract}

\myfootnote{*}{The authors contribute equally to this paper}
\myfootnote{\ddagger}{Currently at Google}

\section{Introduction}
\label{sec:Introduction}

3D object detection aims to recognize and localize objects in a 3D scene by specifying them with their oriented bounding boxes and semantic classes.
Compared to 2D images, 3D scenes provide rich geometric structure information and hence are crucial for many advanced 3D vision applications such as autonomous driving, AR/VR, and robot navigation.
Recent research efforts~\cite{qi2019deep, shi2019pointrcnn, yang20203dssd, zhou2018voxelnet, shi2020points, s18103337, choy20194d, shi2020pv} have been made on 3D object detection and achieve significant progress.
%
However, most existing methods are data-hungry and rely on large-scale labeled 3D objects, leading to a vast amount of costly manual efforts.
To address this issue, it is favorable to develop semi-supervised learning (SSL) algorithms for 3D object detection where plenty of unlabeled 3D point clouds can be leveraged to compensate for the lack of labeled data and to improve detector training.

Several SSL approaches~\cite{laine2016temporal, tarvainen2017mean, sohn2020fixmatch, sohn2020simple, liu2021unbiased} for 2D object detection are developed based on teacher-student mutual learning, where pseudo-labels of unlabeled data are estimated and used as supervisory signals for detector training.
For 3D object detection, 3DIoUMatch~\cite{wang20213dioumatch} employs two identical pre-trained networks to initialize a teacher-student model and applies asymmetric data augmentations to transform data samples.
To be specific, the input data to the student model are globally transformed by strong augmentations for data regularization and variance enhancement, thus offering rich information to boost the capability of the student model.
On the other hand, the input data to the teacher model are obtained by weak augmentations to generate pseudo-labels to supervise the student model.
Prior work~\cite{berthelot2019mixmatch, sohn2020simple, liu2021unbiased, zhao2020sess, wang20213dioumatch} shows that this asymmetric data augmentation mechanism is crucial for improving semi-supervised learning in a teacher-student model.
However, most existing SSL methods for 3D object detection, such as 3DIoUMatch~\cite{wang20213dioumatch}, adopt scene-level transformations, which is sub-optimal as augmenting irrelevant backgrounds may degrade the effectiveness of the augmented data.
To address this issue, we present a method that takes both global and object-level data augmentations into consideration and thus generates more plausible augmented point cloud objects for SSL.

Compared to 3DIoUMatch which applies augmentations such as rotation and scaling to the entire point cloud {\em scene}, our method focuses on point cloud {\em object} augmentation, which better benefits the teacher-student framework.
In this work, we present {\ourname} based on a teacher-student mutual learning framework with an object-level augmentor for semi-supervised 3D object detection.
To this end, we utilize a two-stage training procedure, including the pre-training and semi-supervised learning stages.
First, we design an adversarial formulation to jointly pre-train a detector with an augmentor, where the augmentor takes point clouds within the object bounding box as the input, as well as the objectness guidance from the detector to control the learning pace in augmentation.
Then, the augmentor outputs displacement values for each point as augmentation to improve data variations for the detector.

In the semi-supervised learning stage, we freeze the learned augmentor and use it to produce the object-level augmented point clouds.
We leverage both ground-truth and pseudo-labeled bounding boxes inferred by the teacher model, respectively from the labeled and unlabeled data, to identify point cloud objects that serve as the input to the augmentor.
As a result, the produced point clouds exhibit local variations and are complementary to those produced by global scene-level augmentations, thus improving the teacher-student model learning.
In experiments, we show that our {\ourname} performs favorably against the state-of-the-art methods for semi-supervised object detection on two benchmark datasets, including ScanNet~\cite{Dai_2017_CVPR} and SUN RGB-D~\cite{Song_2015_CVPR}.
In addition, we demonstrate that the proposed augmentor is effective when it is applied to labeled or unlabeled point clouds, and is beneficial from our designed augmentor loss function that is aware of the objectness score from the detector.
The main contributions of this work are summarized as follows:
\begin{enumerate}

\vspace{-2mm}
\item We propose a simple yet effective method for semi-supervised 3D object detection via introducing an object-level augmentation strategy in point cloud scenes.

\vspace{-2mm}
\item We integrate the proposed augmentor into the teacher-student mutual learning framework and jointly train the entire model to make use of labeled and unlabeled data.

\vspace{-2mm}
\item We design a learning mechanism to make augmentor aware of the objectness from the detector and thus generate appropriate augmentations to improve 3D object detection.

\end{enumerate}
\section{Related Work}
\label{sec:RelatedWork}

\paragraph{Semi-supervised Learning.}
Semi-supervised learning (SSL) aims to train a model using few labeled data and abundant unlabeled data.
Numerous SSL strategies have been developed in the literature. 
1) {\em Consistency regularization}: Methods of this category such as~\cite{sajjadi2016regularization, xie2020unsupervised, berthelot2019mixmatch, laine2016temporal} apply different transformations to a data sample and enforce consistency of model predictions among the transformed samples. 
2) {\em Teacher-student framework}: It often employs two identical networks, one for a teacher model and the other for a student model~\cite{tarvainen2017mean, sohn2020fixmatch}. 
The teacher model is first frozen to guide the student model and is then updated from the student model.
3) {\em Pseudo-labeling}: It usually works in a self-supervised manner and derives the model using unlabeled data with their estimated pseudo-labels~\cite{lee2013pseudo}.
Fixmatch~\cite{sohn2020fixmatch} combines the teacher-student framework and pseudo-labeling.
It utilizes both student's and the teacher's predictions to enhance the quality of pseudo-labels.
One key component of this method is asymmetric data augmentation.
The strongly augmented inputs, e.g., those via Mixup~\cite{zhang2017mixup}, to the student model enrich data variance for model training,
while the weakly augmented inputs to the teacher model ensure more accurate pseudo-labels for supervision.
Based on the teacher-student framework, we propose an effective object-level augmentation method that focuses on point cloud instances in a scene.

\vspace{-4mm}
\paragraph{Semi-supervised Object Detection.}
For 2D object detection in SSL, consistency-based methods~\cite{NEURIPS2019_d0f4dae8, tang2021proposal, tang2021humble} enforce the prediction consensus over different augmentations.
Moreover, self-supervised approaches~\cite{sohn2020simple, li2020improving, Wang_2018_CVPR} apply a teacher-student framework with pseudo-label supervisions~\cite{sohn2020simple, zhou2021instant, xu2021end, li2021rethinking, liu2021unbiased, tang2021humble}.
For instance, STAC~\cite{sohn2020simple} and Unbiased Teacher~\cite{liu2021unbiased} apply the teacher-student framework with asymmetric data augmentation to enlarge data variance and filter pseudo-labels to keep high-confidence object proposals.
However, for the 3D scenario, there are fewer explorations of SSL for 3D object detection.
SESS~\cite{zhao2020sess} enforces consistency over different augmentations as regularization.
Furthermore, 3DIoUMatch~\cite{wang20213dioumatch} designs a 3D IoU estimation module based on VoteNet~\cite{qi2019deep} as an IoU-aware Votenet, which calculates the IoU score of object proposals.
Then, it takes IoU scores into account to filter out low-confidence pseudo-labels, with a selective mechanism to supervise unlabeled data using filtered high-quality pseudo-labels.
In contrast, the proposed {\ourname} introduces object-level point augmentations, which is an essential step towards a successful teacher-student framework for SSL, and has not been widely studied in 3D object detection.

\vspace{-4mm}
\paragraph{Data Augmentation on Point Clouds.}
Data augmentation is important for deep learning.
Because training data cannot cover all kinds of scenarios in the complex world, data augmentation is utilized to enlarge the diversity of training data.
In 3D point cloud tasks, global augmentation operations like rotation, scaling, and translation with point-wise jittering~\cite{qi2017pointnet, qi2017pointnet++} are commonly used. 
However, those augmentation methods cannot transform the local structure in a point cloud.
Therefore, recent works aim to improve the augmentation strategies for point clouds.
The method in~\cite{choi2020part} divides an object and applies different augmented operations in each partition.
Moreover, PointAugment~\cite{li2020pointaugment} trains an auto-augmentor network that can learn to augment point cloud samples for better point cloud classification.
PointWOLF~\cite{Kim_2021_ICCV} presents another method for the classification task where a convex combination of multiple transformations with smoothly varying weights carries out the local structure augmentation.
Based on the Mixup~\cite{zhang2017mixup} idea in images, PointMixup~\cite{chen2020pointmixup} interpolates two point cloud objects to create an augmented point cloud, and the model is trained to predict the ratio of two mixed classes with a soft label.
%
PointMixSwap~\cite{umam2022point} further explores the structural variance across multiple point clouds and generates more diverse point clouds for training data enrichment.
%
For 3D object detection, PPBA~\cite{cheng2020improving} iteratively finds the best augmentation parameters of specific operations and applies them to the entire scene.

Compared to the above-mentioned methods that focus on the classification task or the combination of pre-defined augmentation operations, we study the SSL setting for 3D object detection by introducing a simple yet effective augmentation method.
We focus on learning an augmentor that can synthesize object-level point clouds for foreground objects, serving as a better asymmetric augmentation module that is jointly trained in a teacher-student framework to achieve better SSL performance.

\section{Proposed Method}
\label{sec:Method}
This section elaborates the proposed method {\ourname}.
We give the problem definition and method overview in Section \ref{ssec:Method_1}, and then describe our object-level augmentor and its training pipeline in Section \ref{ssec:Method_3}.

\newcolumntype{Y}{>{\centering\arraybackslash}X}

\subsection{Problem Definition and Algorithmic Overview}
\label{ssec:Method_1}

Given a 3D point cloud scene of $S$ points $\mathbf{x} \in\mathbb{R}^{S \times 3}$, 3D object detection aims to recognize and locate objects of interest in $\mathbf{x}$ and describe them by their semantic classes and oriented bounding boxes.
For learning a 3D object detector under the semi-supervised setting, we are given $ N_{l} $ labeled scenes ${\{}\mathbf{x}_{i}^{l}, \mathbf{y}_{i}^{l} {\}}_{i=1}^{N_{l}}$ and $ N_{u} $ unlabeled scenes ${\{}\mathbf{x}_{i}^{u}{\}}_{i=1}^{N_{u}}$, where $ N_{l} << N_{u}$ in practice.
The ground-truth annotation $\mathbf{y}^{l}_{i}$ stores the oriented bounding boxes $\{b_{k}\}$ and semantic labels $\{{c}_{k}\}$ of the objects of interest $\{{o}_{k}\} $ in $\mathbf{x}^{l}_{i}$.

Teacher-student knowledge distillation with asymmetric data augmentation has shown its effectiveness for semi-supervised 3D object detection.
However, previous works~\cite{zhao2020sess,wang20213dioumatch} focus on scene-level augmentation and ignore that object-level variances are crucial for detection.
One way to address this issue is to apply augmentations, e.g., a random rotation, flip, and scale, to the point clouds within each object bounding box.
However, such a method is sub-optimal, and its performance depends on proper augmentation settings.
In Table \ref{Tab:ObservationTable}, we find that pre-defined random augmentations, especially rotations, may confuse model learning and even harm the performance significantly.
%
For scene-level augmentation on 3D object detection, rotation is widely used to enhance data variance without changing geometric relationships between foreground objects and background.
However, for object-level augmentation in a scene, each object has its own orientation with respect to the global scene.
Thus, changing the object-scene context during augmentation may lead to negative effects.
%

%
\begin{table}[t]
\caption{Results of pre-defined object-level augmentations.}
\vspace{-3mm}
\label{Tab:ObservationTable}
\scriptsize
\centering
\renewcommand{\arraystretch}{1.1}
\setlength{\tabcolsep}{1.1pt}
\begin{center}
\begin{tabularx}{1.0\textwidth}{cc*{3}{Y}}
\toprule
\multirow{2}{*}{Setting} & \multicolumn{2}{c}{ScanNet 10\%} & \multicolumn{2}{c}{SUN RGB-D 5\%} \\
\cline{2-5}
                    & mAP@0.25  & mAP@0.5  & mAP@0.25 & mAP@0.5 \\
\hline
Without Object-level Aug.      & 47.1 & 28.3 & 39.0 & 21.1 \\
Pre-defined Object-level Aug. (scale, flip, rotation) & 42.7 & 24.2 & 24.9 & 13.6 \\
Pre-defined Object-level Aug. (displacement, range at 0.5\%) & 48.4 & 29.1 & 40.6 & 20.4 \\
Pre-defined Object-level Aug. (displacement, range at 1\%) & 49.0 & 29.3 & 40.5 & 20.9 \\
Pre-defined Object-level Aug. (displacement, range at 5\%) & 47.3 & 27.4 & 39.5 & 20.5 \\
\bottomrule
\end{tabularx}
\end{center}
\vspace{-5mm}
\end{table}

%
%
We instead consider point displacement for augmentaiton since it can enhance object-level data variance while keeping object orientations.
As shown in Table \ref{Tab:ObservationTable}, we try different ranges of displacement.
%
%
%
Although using random displacements slightly improves the performance, it requires to pre-define a proper range of displacement, e.g., using too large or too small displacements may not be optimal.
These issues motivate us to develop a better strategy via learning an augmentor for object-level augmentation that benefits 3D object detection.
By learning an augmentor to generate proper displacement values, we preserve the intrinsic characteristics of an object and avoid over-deforming it.

\begin{figure*}[!t]
	\centering
	    \includegraphics[width=0.95\linewidth]{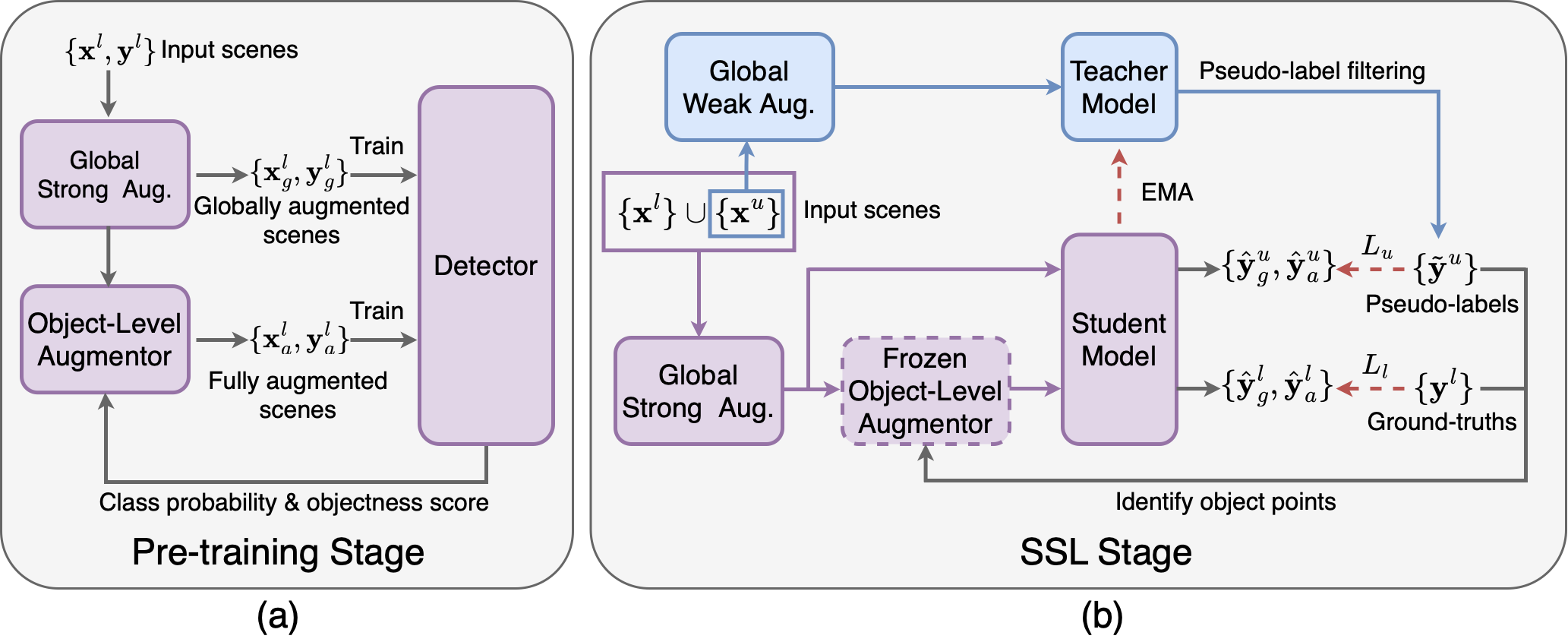}
	\vspace{-3mm}
	\caption{
        \textbf{{\ourname} pipeline at (a) the pre-training stage and (b) the SSL stage.} 
        In the pre-training stage, we utilize globally and fully augmented labeled scenes to jointly train the detector and augmentor using an adversarial strategy.
        In the SSL stage, we leverage the teacher-student framework with our frozen object-level augmentor. The teacher model consumes unlabeled data to generate high-quality pseudo-labels. Both labeled and unlabeled data are globally augmented and fully augmented to train the student model, where the augmentor takes points within each object bounding box as input and outputs the augmented points. Finally, the teacher model is updated from student model via EMA.
        }
	\label{fig:pipeline}
	\vspace{-3mm}
\end{figure*}

\vspace{-3mm}
\paragraph{Teacher-Student Framework in SSL.}
\label{ssec:Method_2}
We aim to learn an augmentor that can synthesize plausible object instances while excluding irrelevant backgrounds, without twisting any augmentation parameters.
Moreover, the augmentor can be integrated into the teacher-student framework and supports SSL.
Fig.~\ref{fig:pipeline}(b) shows the training pipeline.
The teacher and student models are initialized from the same model. 
The teacher model is updated from the student model using the exponential moving average (EMA) mechanism, while pseudo-labels of unlabeled data are generated by the teacher model and are filtered to provide high-quality labels to the student model.
The ground-truth and pseudo-labeled bounding boxes respectively from the labeled and unlabeled data are used to supervise the student model.

A key component making the teacher-student framework effective is data augmentation.
We first follow~\cite{wang20213dioumatch} to apply the global transformations (e.g., rotation, flip, scale) to point cloud scenes, where the weak and strong augmentations are used for the teacher and student models, respectively.
More details can be referred to~\cite{wang20213dioumatch}.
To integrate our object-level augmentor, after global augmentation, we apply our augmentor to points within each object bounding box. Note that we only use the augmentor for the student model (see Fig.~\ref{fig:pipeline}(b)), as the student model is the main model for updating parameters from loss functions.
In practice, we also have tried to apply our augmentor to the teacher model but it does not show significant differences.
To train our augmentor, we utilize a pre-training stage, shown in Fig.~\ref{fig:pipeline}(a), to jointly train the augmentor and detector using only the labeled data. 
The reason is that, in the SSL stage, we find that using the unlabeled data with noisy pseudo-labeled bounding boxes would cause instability in training the augmentor. 
More details are described in the following section.

\subsection{Object-level Point Augmentor}
\label{ssec:Method_3}

In this work, we aim to train an object-level augmentor that can determine point-wise parameters for foreground points and increase the variation of local structure in a scene.
Different from PointAugment~\cite{li2020pointaugment}, we use only the point-wise displacement $\mathbb{D}$ to transform object points since we observe that random rotation is not helpful in 3D object detection as mentioned in Section \ref{ssec:Method_1}.
In addition, we dynamically learn the augmentor that controls the appropriate magnitude of point displacement based on objectness scores of the detector.
Lastly, we leverage both labeled and unlabeled data to mutually update both the detector and the augmentor via an adversarial learning strategy.

\begin{figure*}[!t]
	\centering
        \includegraphics[width=0.95\linewidth]{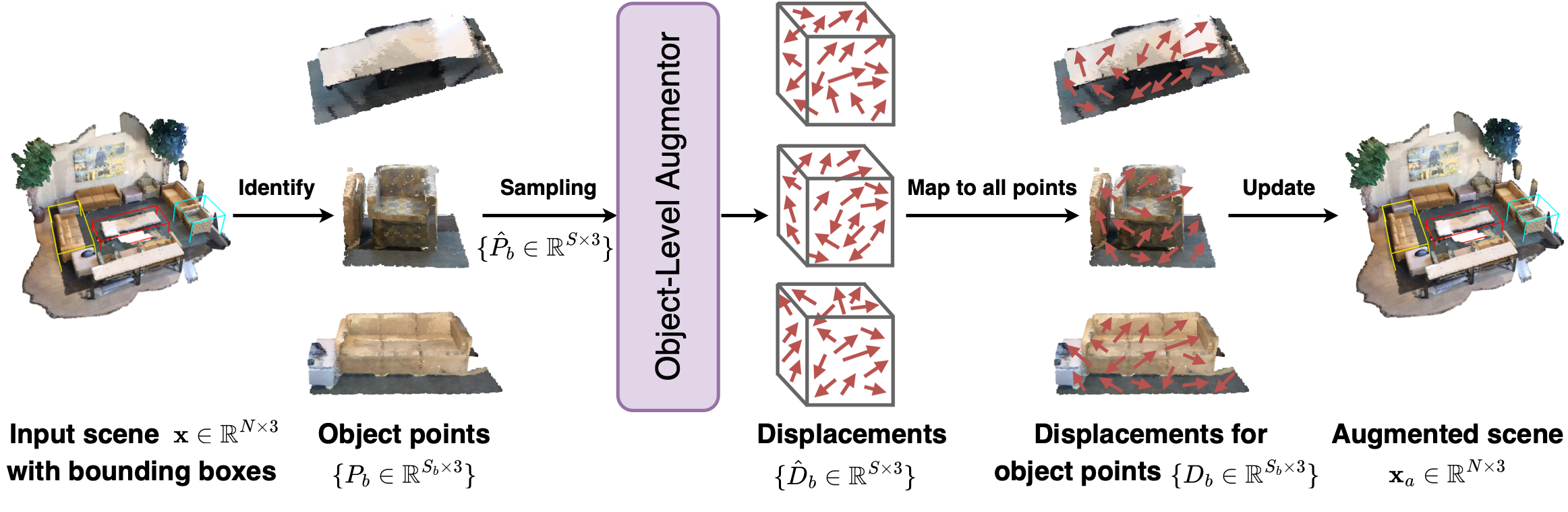}
	\vspace{-2mm}
	\caption{
        Given a point cloud scene $\mathbf{x} \in \mathbb{R}^{N \times 3}$ with $M$ objects, we identify the object points $\{P_{b} \in \mathbb{R}^{S_{b} \times 3}\}_{b=1}^M$ in the $M$ bounding boxes, where $S_{b}$ is the number of points within the $b$th bounding box. 
        Point sampling is applied to $\{P_{b}\}$ and makes each of the resultant sampled objects $\{\hat{P}_{b}\}$ have $S$ points, which then serve as the input to the augmentor.
        The augmentor outputs the displacements $\{\hat{D}_{b} \in \mathbb{R}^{S \times 3}\}$. 
        We map them back to their original szies $\{{D}_{b} \in \mathbb{R}^{S_{b} \times 3}\}$ via reverse sampling. 
        Finally, $\{{D}_{b}\}$ is added back to the scene to obtain the object-level augmented scene ${\mathbf{x}}_{a} \in \mathbb{R}^{N \times 3}$.
        }
	\label{fig:detail}
\end{figure*}

\vspace{-2mm}
{\flushleft \bf {Augmentation Process.}}
The augmentation processing is illustrated in Fig.~\ref{fig:detail}.
Given a globally augmented 3D point cloud scene containing objects and their bounding boxes $\{\mathbf{x}^{l}_{g}, \mathbf{y}^{l}_{g}\}$, we sample $M$ foreground objects from this scene to apply object-level augmentations. 
For unlabeled scene, we utilize its pseudo-labeled bounding boxes $\{\mathbf{x}^{u}_{g}, \tilde{\mathbf{y}}^{u}_{g}\}$.
For each scene, the points inside the $M$ bounding box proposals are collected, i.e., $\{ P_{b} \in \mathbb{R}^{S_b \times 3} \}_{b=1}^{M} $, where $S_b$ is the number of points inside the $b$th proposal.
Then, we either up-sample by padding or down-sample by farthest point sampling (FPS) to make each object have exactly $S$ points $\{ \hat{P}_{b} \in {\mathbb{R}^{S \times 3}} \}_{b=1}^{M} $, while keeping each object structure unchanged.
The augmentor takes the sampled objects $\{ \hat{P}_{b} \}_{b=1}^{M} $ as input and outputs point-wise displacements $\{ \hat{D}_{b} \in {\mathbb{R}^{S \times 3}} \}_{b=1}^{M} $ for point clouds $\{ \hat{P}_{b} \}_{b=1}^{M} $.
To match displacement $\{ \hat{D}_{b} \}_{b=1}^{M}$ back to the point clouds of the original sizes $\{ D_{b} \in {\mathbb{R}^{S_b \times 3}} \}_{b=1}^{M}  $, we record the mapping from $P_b$ to $\hat{P}_{b}$ and apply the reverse mapping.
The point-wise displacement $\{ D_{b} \}_{b=1}^{M}$ is added to the object points $\{ P_{b} \}_{b=1}^{M}$ as our object-level augmentation.
The fully augmented scene $\mathbf{x}_{a}$ is obtained by replacing the original object points with the augmented points.
%
Note that after obtaining augmented points, we restrict the displacement not to exceed the original bounding box.
The original object points within the bounding box can be replaced by the augmented points that fit the original background while not affecting other objects.
%

\vspace{-2mm}
{\flushleft \bf {Joint Augmentor and Detector Training.}}
We use labeled data, including globally augmented samples $\{\mathbf{x}^{l}_{g}\}$ and fully augmented samples $ \{\mathbf{x}^{l}_{a} \}$ via our augmentor, to jointly train the detector and augmentor in the pre-training stage. 
The augmentor is optimized to generate proper augmented scene $\mathbf{x}^{l}_{a}$ and to maximize the detector capability, while the detector is derived to localize and recognize the augmented data accurately.

\vspace{-1mm}
{\flushleft \bf {Detector Loss.}}
For training the detector, we formulate the loss function $ \mathcal{L}_D $ as follows:
\begin{equation}
\mathcal{L}_{D} = \mathcal{L}_d(\mathbf{x}^{l}_{g}, \mathbf{y}^{l}_g) + \mathcal{L}_d(\mathbf{x}^{l}_{a}, \mathbf{y}^{l}_a), 
\label{eqn:detectorLoss}
\end{equation}
where $\mathcal{L}_d$ is the detection loss used in~\cite{wang20213dioumatch}.

\vspace{-1mm}
{\flushleft \bf {Augmentor Loss.}} 
Similar to PointAugment~\cite{li2020pointaugment}, the fully augmented sample $\mathbf{x}^{l}_{a}$ should satisfy the following two requirements:
1) Predicting $\mathbf{x}^{l}_{a}$ should be more challenging than $\mathbf{x}^{l}_{g}$, i.e., $ \mathcal{L}_d(\mathbf{x}^{l}_{a}, \mathbf{y}^{l}_a) \geq \mathcal{L}_d(\mathbf{x}^{l}_{g}, \mathbf{y}^{l}_g)$;
2) ${\mathbf{x}}^{l}_{a}$ and ${\mathbf{x}}^{l}_{g}$ should be similar to some degree by enforcing that they are predicted as the same class.
To satisfy the two requirements, we use a dynamic variable $\rho$ to control the augmentation magnitude.
%
%
$\mathcal{L}_d(\mathbf{x}^{l}_{a}, \mathbf{y}^{l}_a)$ should be larger than $\mathcal{L}_d(\mathbf{x}^{l}_{g}, \mathbf{y}^{l}_g)$ for the first requirement and should not become too far for the second requirement.
Thus, we make $\rho \mathcal{L}_d(\mathbf{x}^{l}_{g}, \mathbf{y}^{l}_g)$ be the upper bound of $\mathcal{L}_d(\mathbf{x}^{l}_{a}, \mathbf{y}^{l}_a)$.
%
%
With a larger value of $\rho$, the augmentor generates more challenging augmented samples.
On the other hand, the smaller value of $\rho$ can avoid over-deforming the augmented samples.
The augmentor loss $\mathcal{L}_{A}$ is formulated as 
\begin{equation}
\mathcal{L}_{A} = \mathcal{L}_d(\mathbf{x}^{l}_{a}, \mathbf{y}^{l}_{a}) + \lambda \lvert 1 - \exp{(\mathcal{L}_d(\mathbf{x}^{l}_{a}, \mathbf{y}^{l}_a) - \rho \mathcal{L}_d(\mathbf{x}^{l}_{g}, \mathbf{y}^{l}_g))} \rvert ,  
\label{eqn:augmentorLoss}
\end{equation}
where $\lambda$ is a pre-defined constant used to balance the importance between the object detection term and the augmentation magnitude term.
$\rho \geq 1$ is set to ensure $ \mathcal{L}_d(\mathbf{x}^{l}_{a}, \mathbf{y}^{l}_a) \geq \mathcal{L}_d(\mathbf{x}^{l}_{g}, \mathbf{y}^{l}_g)$, while $\rho$ cannot be too high otherwise the augmented samples become too challenging. 
To balance it, we follow \cite{li2020pointaugment} and bound $\rho$ between 1 and a value based on the classification probability.
Different from \cite{li2020pointaugment}, we further include a term $\hat{\mathbf{y}}_{o}$ to make $\rho$ aware of the objectness for our object detection task:
\begin{equation}
		\rho = \operatorname{max}(1, \exp( \hat{\mathbf{y}}_{o} \cdot \sum_{c=1}^{C} \hat{\mathbf{y}}_{c} \cdot {\mathbf{y}}_{c} )),
\label{eqn:rho}
\end{equation} 
where $C$ is the number of classes. ${\mathbf{y}}_{c}$, $\hat{\mathbf{y}}_{c}$, and $\hat{\mathbf{y}}_{o}$ are the class label, classification probability, and objectness score, respectively.

We find that our introduced $\hat{\mathbf{y}}_{o}$ term is critical to our task.
%
As a metric to evaluate the objectness ability, $\hat{\mathbf{y}}_{o}$ is more suitable than the IoU score which is too sensitive to the bounding box location.
%
%
When the class probability or the objectness score, \ie $\hat{\mathbf{y}}_{o}$, of a sample is higher, it implies that this sample can be well classified by the detector, so we may use a larger value of $\rho$ to allow more augmentations and make the augmented sample more challenging.
%
%
%
%
Since the augmentor is learned in a class-agnostic fashion, the objectness score provides class-agnostic guidance to control the difficulty of the augmented samples, which in turn improves the learning of object detector.
Finally, we alternatively train $\mathcal{L}_D$ and $\mathcal{L}_A$ in the pre-training stage.

\vspace{-1mm}
{\flushleft \bf {Overall Loss Functions for SSL.}}
In the SSL stage, we initialize the student and teacher models from the pre-trained detector and freeze the augmentor.
The training pipeline is illustrated in Fig.~\ref{fig:pipeline}(b).
In each training batch, there are labeled samples $\{ \mathbf{x}^{l}, \mathbf{y}^{l} \}$ and unlabeled samples $\{ \mathbf{x}^{u} \}$.
After applying global augmentation and our augmentor, we collect four kinds of data for the student model to learn: globally augmented labeled data $\{\mathbf{x}^{l}_{g}\}$ and unlabeled data $\{\mathbf{x}^{u}_{g}\}$, fully augmented labeled data $\{\mathbf{x}^{l}_{a}\}$ and unlabeled data $\{\mathbf{x}^{u}_{a}\}$.
The student model outputs corresponding predictions: $\hat{\mathbf{y}}^{l}_{g}$, $\hat{\mathbf{y}}^{u}_{g}$, $\hat{\mathbf{y}}^{l}_{a}$, and $\hat{\mathbf{y}}^{u}_{a}$. 
For labeled data, $\hat{\mathbf{y}}^{l}_{g}$ and $\hat{\mathbf{y}}^{l}_{a}$ are supervised with the ground truths via
\begin{equation}
\label{eqn:SSL_L}
	\mathcal{L}_{l} = 
    \mathcal{L}_d(\mathbf{x}^{l}_{g}, \mathbf{y}^{l}_g) + 
    \mathcal{L}_d(\mathbf{x}^{l}_{a}, \mathbf{y}^{l}_a).
\end{equation}
For unlabeled data, $\hat{\mathbf{y}}^{u}_{g}$ and $\hat{\mathbf{y}}^{u}_{a}$ are supervised by filtered pseudo-labels $\tilde{\mathbf{y}}^{u}$:
\begin{equation}
\label{eqn:SSL_U}
	\mathcal{L}_{u} = 
    \mathcal{L}_d(\mathbf{x}^{u}_{g}, \tilde{\mathbf{y}}^{u}_g) +
    \mathcal{L}_d(\mathbf{x}^{u}_{a}, \tilde{\mathbf{y}}^{u}_a). 
\end{equation}
The overall loss in SSL for both labeled and unlabeled data is $\mathcal{L}_{SSL} = \mathcal{L}_{l} + \mathcal{L}_{u}$.
%
The teacher model is updated by Exponential Moving Average (EMA) from the student model.


\section{Experiments}

\paragraph{Datasets.}
We follow the settings in the prior work~\cite{wang20213dioumatch, zhao2020sess} for semi-supervised 3D object detection. 
ScanNet~\cite{Dai_2017_CVPR} is a 3D indoor benchmark dataset. 
It contains 1,201 training and 312 validation scenes with the reconstructed meshes.
We focus on the 18 semantic classes.
SUN RGB-D~\cite{Song_2015_CVPR} is another 3D indoor benchmark dataset. 
It is composed of 5,285 training and 5,050 validation scenes. 
We use 10 object classes to evaluate our model.

\vspace{-3mm}
\paragraph{Evaluation Metrics.} For both benchmarks, we split them into the labeled and unlabeled data to perform semi-supervised learning. We apply 5\%, 10\%, and 20\% labeled data ratio settings to conduct our experiments.
We adopt mAP (mean average precision) as the evaluate metrics and report mAP@0.25 (mAP with 3D IoU threshold at 0.25) and mAP@0.5 scores.

\vspace{-3mm}
\paragraph{Implementation Details.} For pre-training, we use a batch size as 4 to train the augmentor. We use $M=3$ foreground objects in one scene and sample $S=1024$ points using either FPS or point padding according to the original point size. 
We train the detector and the augmentor for 900 epochs and use the Adam optimizer with an initial learning rate of 0.001. The learning rate decay by 0.1 occurs in the $400^{th}$, $600^{th}$, and $800^{th}$ epoch. To further stabilize the training, we leverage a warm-up mechanism that does not train the augmentor for the first 100 epochs.
For the augmentor loss \eqref{eqn:augmentorLoss}, we set $\lambda = 0.1$.

In the SSL stage, a batch is composed of two labeled data and four unlabeled data.
We leverage ground truth bounding boxes to identify $S = 3$ foreground objects in labeled data, while for unlabeled data, we randomly pick $S = 3$ foreground objects from top-6 pseudo-labels with the highest confidence calculated by the IoU and objectness scores from the detector outputs.
The IoU score represents the localization quality of the proposals and the objectness score shows the classification quality. We take both them into account and select the pseudo-labels of high quality.
This mechanism avoids some easy samples with high confidence being selected all the time, which increases the chance that the model can observe more data variations.
We train the detector for 1,000 epochs and use the Adam optimizer with an initial learning rate of 0.002. The learning rate decays 0.3, 0.3, 0.1, 0.1 at $400$th, $600$th, $800$th, and $900$th epochs, respectively.
We conduct experiments on a single GTX 2080-Ti GPU.
For fair comparisons, we follow the procedure in \cite{wang20213dioumatch} to use the student model for inference, along with a post-processing step on final predictions.

\begin{table}[!t]
\caption{
    Results on ScanNet val set and SUN RGB-D val set for 5\%, 10\%, 20\% labeled data ratio.
    We run the experiments under 3 random data splits and report our result in mean±standard deviation for the mAP@0.25 and mAP@0.50 metric.
    }
\vspace{-1mm}
\label{Tab:SotaTable}
\scriptsize
\centering
\renewcommand{\arraystretch}{1.2}
\setlength{\tabcolsep}{1.1pt}
\begin{center}
\begin{tabularx}{\textwidth}{cc*{6}{Y}}
\toprule
\multirow{3}{*}{Dataset} & \multirow{3}{*}{Model} & \multicolumn{2}{c}{5\%} & \multicolumn{2}{c}{10\%} & \multicolumn{2}{c}{20\%} \\
\cline{3-8}
& & mAP & mAP & mAP & mAP & mAP & mAP \\
& & @0.25 & @0.5 & @0.25 & @0.5 & @0.25 & @0.5 \\
\hline
\multirow{5}{*}{ScanNet} & VoteNet~\cite{qi2019deep} & 27.9±0.5 & 10.8±0.6 & 36.9±1.6 & 18.2±1.0 & 46.9±1.9 & 27.5±1.2 \\
& SESS~\cite{zhao2020sess} & NA & NA & 39.7±0.9 & 18.6 & 47.9±0.4 & 26.9 \\
& 3DIoUMatch~\cite{wang20213dioumatch} & 40.0±0.9 & 22.5±0.5 & 47.2±0.4 & 28.3±1.5 & 52.8±1.2 & 35.2±1.1 \\
& {\ourname} & \textbf{41.9}±1.5 & \textbf{25.0}±0.4 & \textbf{50.5}±0.2 & \textbf{32.7}±1.0 & \textbf{54.7}±0.3 & \textbf{36.8}±0.8 \\
& Gain (\%) & 1.9↑ & 2.5↑ & 3.3↑ & 4.4↑ & 1.9↑ & 1.6↑ \\
\hline
\multirow{5}{*}{SUN RGB-D} & VoteNet~\cite{qi2019deep} & 29.9±1.5 & 10.5±0.5 & 38.9±0.8 & 17.2±1.3 & 45.7±0.6 & 22.5±0.8 \\
& SESS~\cite{zhao2020sess} & NA & NA & 42.9±1.0 & 14.4 & 47.9±0.5 & 20.6 \\
& 3DIoUMatch~\cite{wang20213dioumatch} &39.0±1.9 & 21.1±1.7 & 45.5±1.5 & 28.8±0.7 & 49.7±0.4 & 30.9±0.2 \\
& {\ourname} & \textbf{41.6}±0.1 & \textbf{23.1}±0.5 & \textbf{47.2}±0.7 & \textbf{29.6}±0.8 & \textbf{50.8}±1.0 & \textbf{31.5}±0.6 \\
& Gain (\%) & 2.6↑ & 2.0↑ & 1.7↑ & 0.8↑ & 1.1↑ & 0.6↑ \\
\bottomrule
\end{tabularx}
\end{center}
\vspace{-5mm}
\end{table}

\subsection{Experimental Results}

\subsubsection{Main Results}
Table~\ref{Tab:SotaTable} shows the result of our method on ScanNet and SUN RGB-D, under different labeled data ratios compared with state-of-the-art methods for 3D object detection in SSL, including VoteNet~\cite{qi2019deep}, SESS~\cite{zhao2020sess}, and 3DIoUMatch~\cite{wang20213dioumatch}. 
The proposed OPA method consistently performs favorably against existing approaches in all the settings.
Moreover, our method performs better in settings with lower labeled data ratios, e.g.,  SUN RGB-D 5\% and  ScanNet 10\%, which shows the advantage of the proposed augmentor.
Note that, since the total number of scenes in ScanNet is five times less than the one in SUN RGB-D, we find that the performance gain of 5\% ScanNet is slightly less than the 10\% ScanNet setting, which can be caused by the less data to train the augmentor.
More results and analysis are provided in the supplementary material.

\subsubsection{Ablation Study}
\paragraph{Augmentation on Labeled and Unlabeled Data.}
We first study the effect of our augmentor trained on labeled or unlabeled data. In Table~\ref{Tab:AblationTable}, comparing to ID (5) using our augmentor on both labeled and unlabeled data (i.e., our full model), we show the benefit by removing labeled or unlabeled data in ID (2) and (3), respectively.
%
%
Moreover, comparing ID (1) with ID (2) and ID (3), where we include unlabeled and labeled data in our proposed augmentor with $\hat{y}_{o}$, the performance gains (ScanNet 10\% mAP@0.5) are 3.1\% and 3.4\%, respectively.
In ID (2), the augmentor helps unlabeled data to produce better data variance for student model training.
%
In ID (3), the augmentor provides more diverse supervised samples in the pre-training stage.
%
This shows that our augmentor can take advantage of different data and improve performance.
Note that, experiments are conducted in one of the same data splits for fair comparisons, and thus the numbers of our full model are slightly different from the averaged numbers in Table~\ref{Tab:SotaTable}.

\vspace{-3mm} 
\paragraph{Objectness Term $\hat{y}_o$ in \eqref{eqn:rho}.}
Different from PointAugment~\cite{li2020pointaugment}, we introduce an objectness term in~\eqref{eqn:rho} that controls the magnitude of augmentation, so that the augmentor is aware of the quality of class-agnostic detection results and learns how to generate appropriate augmentations with challenging variations.
In Table~\ref{Tab:AblationTable}, ID (4) without using this objectness term performs worse than our full model in ID (5), which indicates that this term is essential to generate augmentations that are helpful in our SSL setting.

\begin{table}[!t]
\caption{
    We study the affect of proposed components in our augmentor in settings of ScanNet 10\% and SUN RGB-D 5\% labeled data ratio.
    }
\vspace{-2mm}
\label{Tab:AblationTable}
\scriptsize
\centering
\renewcommand{\arraystretch}{1.2}
\begin{center}
\begin{tabularx}{0.95\textwidth}{cc*{6}{Y}}
\toprule
\multirow{3}{*}{ID} &
\multirow{3}{*}{Aug. (labeled)} & \multirow{3}{*}{Aug. (unlabeled)} & \multirow{3}{*}{ $\hat{y}_{o}$ in \eqref{eqn:rho}} & \multicolumn{2}{c}{ScanNet 10\%} & \multicolumn{2}{c}{SUN RGB-D 5\%} \\
\cline{5-8}
& & & & mAP & mAP & mAP & mAP \\
& & & & @0.25 & @0.5 & @0.25 & @0.5 \\
\hline
(1) & &  &  & 47.1 & 28.3 & 38.1 & 21.3\\
(2) & & $\surd$ & $\surd$ & 50.4 & 31.4 & 40.1 & 23.2\\
(3) & $\surd$ &  & $\surd$ & 50.4 & 31.7 & 40.1 & 22.5\\
(4) & $\surd$ & $\surd$ &  & 48.5 & 29.3 & 38.1 & 22.1\\
(5) & $\surd$ & $\surd$ & $\surd$ & 50.7 & 32.4 & 41.8 & 23.5\\
\bottomrule
\end{tabularx}
\end{center}
\vspace{-3mm}
\end{table}

\begin{table}
    \caption{Sensitivity analysis of $\lambda$ in \eqref{eqn:augmentorLoss} on ScanNet 10\% labeled data ratio.}
    \vspace{-2mm}
    \label{Tab:SensitivityTable}
    \scriptsize
    \centering
    \renewcommand{\arraystretch}{1.2}
    \begin{center}
        \begin{tabularx}{0.5\textwidth}{c*{2}{Y}}
            \toprule
            \multirow{2}{*}{$\lambda$ in \eqref{eqn:augmentorLoss}} & \multicolumn{2}{c}{ScanNet 10\%} \\
            \cline{2-3}
            & mAP@0.25 & mAP@0.5 \\
            \hline
            0.01 & 49.8 & 32.1 \\
            0.05 & 50.5 & 31.4 \\
            0.1 & 50.7 & 32.4 \\
            0.5 & 50.1 & 31.7 \\
            1.0 & 48.9 & 29.5 \\
        \bottomrule
        \end{tabularx}
    \end{center}
    \vspace{-7mm}
\end{table}

\vspace{-3mm}
\paragraph{Sensitivity on $\lambda$ in \eqref{eqn:augmentorLoss}.}
In Table~\ref{Tab:SensitivityTable}, we test the sensitivity on $\lambda$ in \eqref{eqn:augmentorLoss} when training the augmentor using 10\% labeled data on ScanNet.
The higher lambda values (e.g., 1.0) accelerate the training processing of our augmentor to become more aggressive (i.e., generating more challenging samples), which may harm the stability in the early training stage, thus leading to worse performance. On the other hand, the lower lambda values control the pace for training the augmentor in an appropriate step, stabilizing the training and leading to better performance.
Overall, Table~\ref{Tab:SensitivityTable} shows that our method is robust to the $\lambda$ value when it is in a reasonable range (e.g., from $0.01$ to $0.5$).
In all the experiments, we choose $\lambda = 0.1$.


\section{Conclusions}
\label{sec:Conclusion}
In this paper, we propose {\ourname}, a novel teacher-student mutual learning framework with object-level augmentor, which benefits semi-supervised learning on both labeled and unlabeled data for 3D object detection.
We show that the existing methods using only global transformations is sub-optimal, and  thus we propose to adopt both global and local augmentations.
To this end, we propose to learn an object-level augmentor that is jointly trained with the object detector in an adversarial learning manner, in which the objectness score from the detector provides the guidance to the augmentor.
In this way, our object-level augmentor is able to increase the variance within object points and thus boost the detector's capability in SSL.
We conduct extensive experiments on the ScanNet and SUN RGB-D benchmarks, in which {\ourname} achieves consistent performance gains against state-of-the-art approaches on all the settings with different ratios of labeled data.

\vspace{-3mm}
\paragraph{Acknowledgment.}
This work was supported in part by National Science and Technology Council (NSTC) under grants 111-2628-E-A49-025-MY3, 109-2221-E-009-113-MY3, 110-2634-F-006-022, 110-2634-F-002-050, and 111-2634-F-007-002. This work was funded in part by Qualcomm and MediaTek.


\section{Supplementary Material}
\label{sec:supp}

We provide more details and analysis of our proposed method, Object-level Point Augmentor ({\ourname}), and is arranged as follows:
We elaborate the experiments using pre-defined augmentations in Section~\ref{ssec:predefined}.
The per-class mAP scores of the detector trained with {\ourname} are reported in Section~\ref{ssec:perclass}.
We evaluate the performance of the pre-trained detector with the proposed augmentor in Section~\ref{ssec:pretrained}.
We analyze the displacement distribution in the form of histograms in Sec.~\ref{ssec:dis}.
Finally, the qualitative visualizations are shown in Section~\ref{ssec:vis}.

\subsection{Details of Pre-defined Augmentation Experiments}
\label{ssec:predefined}
Table 1 of the main submission reports the results of using pre-defined object-level augmentations.
A more detailed description of this experiment is given below.
We test the object-level augmentation with 1) three operations, including scale, flip, and rotation, and 2) only displacement.
For using the three operations, we follow the same augmentation operations that are used in scene-level augmentations but only apply them to the foreground points with a smaller magnitude.
Specifically, we randomly scale an object between 0.85 and 1.15 times the original size, randomly rotate the object from -5 to 5 degrees, and randomly horizontal flip the object with the probability of 0.5.
%
For the experiment using only displacement under a certain range $\alpha$, we randomly jitter each point in the x-direction, y-direction, and z-direction with a $-\alpha\%$ to $\alpha\%$ ratio of displacement with respect to its corresponding bounding box size.

\subsection{Per-class Evaluation}
\label{ssec:perclass}
We report the per-class average precision on the val set of ScanNet~\cite{Dai_2017_CVPR} with 10\% labeled data and of SUN RGB-D~\cite{Song_2015_CVPR} with 5\% labeled data in Table~\ref{Table:preclass-scannet-table} and Table~\ref{Table:preclass-sunrgbd-table}, respectively.
The results in Table~\ref{Table:preclass-scannet-table} show that our method improves the performance in almost all the classes.
Similarly, Table~\ref{Table:preclass-sunrgbd-table} indicates that our model achieves more favorable results in most classes.
Overall, our method achieves better performance against the 3DIoUMatch~\cite{wang20213dioumatch}, the best competing method.

\begin{table}[h]
\caption{Per-class mAP@0.25 (top group) and mAP@0.5 (bottom group) on the ScanNet val set with 10\% labeled data.}
\vspace{2mm}
\centering
\renewcommand{\arraystretch}{1.2}
\resizebox{\columnwidth}{!}{%
\begin{tabular}{ccccccccccccccccccc}
\hline
                    & cabin  & bed  & chair & sofa & table & door & wind & bkshf & pic & cntr & desk & curt & refrig & showr & toilet & sink & bath & ofurn \\ \hline
VoteNet~\cite{qi2019deep}    & 17.9 & 74.7 & 74.5  & 75.3 & 45.6  & 18.3 & 11.7 & 21.7  & 0.7 & 28.4   & 49.4 & 21.5 & 23.2   & 18.5 & 79.6 & 25.7 & 66.3 & 11.7 \\
SESS~\cite{zhao2020sess}       & 20.5 & 75.1 & 76.2  & 76.4 & 48.1  & 20.0 & 14.4 & 19.4  & 1.2 & 30.0   & 51.8 & 25.0 & 30.0   & 26.4 & 82.2 & 29.2 & 72.3 & 14.1 \\
3DIoUMatch~\cite{wang20213dioumatch} & 26.6 & 82.6 & 80.9  & \textbf{83.3} & 52.1  & 28.0 & 19.9 & 29.4  & 3.7 & 45.0   & \textbf{61.9} & 29.2 & 34.1   & 51.2 & 85.7 & 32.3 & 82.8 & 21.5 \\
OPA        & \textbf{27.4} & \textbf{85.7} & \textbf{81.8}  & 79.6 & \textbf{53.8}  & \textbf{32.4} & \textbf{27.6} & \textbf{37.6}  & \textbf{5.3} & \textbf{53.5}   & 58.2 & \textbf{35.7} & \textbf{35.2}   & \textbf{57.1} & \textbf{94.4} & \textbf{33.3} & \textbf{86.3} & \textbf{26.1}   \\ 
\hline
VoteNet~\cite{qi2019deep}     & 3.2 & 64.6 & 43.4  & 49.3 & 25.1  & 2.8 & 1.1 & 8.7  & 0.0 & 2.4   & 14.7 & 3.9 & 7.6   & 1.1 & 46.8 & 11.9 & 39.4 & 1.5 \\
SESS~\cite{zhao2020sess}       & 3.7 & 61.2 & 48.0  & 44.8 & 29.5  & 3.2 & 2.8 & 8.4  & 0.2 & 7.5   & 19.2 & 5.0 & 12.2  & 1.8 & 48.7 & 15.3 & 40.8 & 2.2 \\
3DIoUMatch~\cite{wang20213dioumatch}  & 5.9  & 72.0 & 60.5  & 56.6 & 39.7  & 10.3 & 5.2  & 18.1  & \textbf{0.7} & 16.0   & \textbf{35.3} & 8.3  & 21.4   & 6.2  & 67.5 & 13.2 & 67.6 & 5.2 \\
OPA        & \textbf{6.6}  & \textbf{72.8} & \textbf{64.2}  & \textbf{66.2} & \textbf{41.2}  & \textbf{10.7} & \textbf{9.7}  & \textbf{23.4}  & 0.1 & \textbf{20.0}   & \textbf{35.3} & \textbf{16.2} & \textbf{23.2}   & \textbf{15.9} & \textbf{90.4} & \textbf{20.4} & \textbf{83.1} & \textbf{11.7}   \\ \hline
\end{tabular}
}
\label{Table:preclass-scannet-table}
\vspace{2mm}
\end{table}

\begin{table}
\caption{Per-class mAP@0.25 (top group) and mAP@0.5 (bottom group) on the SUNRGB-D val set with 5\% labeled data.}
\vspace{2mm}
\label{Tab:perclassSUNRGBD}
\centering
\renewcommand{\arraystretch}{1.2}
\resizebox{\columnwidth}{!}{%
\begin{tabular}{ccccccccccc}
\hline
                    & bathtub & bed  & bookshelf & chair & desk & dresser & nightstand & sofa & table & toilet \\ 
\hline
VoteNet~\cite{qi2019deep}    & 67.8    & 32.2 & 39.4      & 58.5  & 53.5 & 8.0     & 1.9       & 14.7 & 3.2   & 20.3   \\
SESS~\cite{zhao2020sess}       & 70.8    & 34.7 & 41.9      & 60.4  & 63.0 & \textbf{9.8}     & 3.7       & 25.2 & 4.0   & 28.0   \\
3DIoUMatch~\cite{wang20213dioumatch} & 75.4    & 37.7 & 45.2      & \textbf{64.2}  & 77.0 & 6.0     & \textbf{5.7}       & 34.6 & \textbf{4.5}   & 39.4   \\
OPA        & \textbf{77.1}    & \textbf{39.6} & \textbf{47.7}      & 63.4  & \textbf{81.1} & 8.2     & 4.8       & \textbf{44.7} & 3.6   & \textbf{49.2}   \\
\hline
VoteNet~\cite{qi2019deep}     & 31.2    & 6.2  & 15.5      & 29.6  & 14.6 & 0.5     & 0.2       & 2.0  & 0.3   & 5.2    \\
SESS~\cite{zhao2020sess}        & 36.7    & 7.2  & 19.2      & 31.8  & 20.4 & 0.7     & 0.5       & 7.0  & 0.4   & 7.1    \\
3DIoUMatch~\cite{wang20213dioumatch}  & 45.2    & \textbf{14.4} & 27.8  & \textbf{43.6}  & 47.2 & 0.8     & \textbf{1.9}       & 15.7 & \textbf{0.6}   & 13.4   \\ 
OPA         & \textbf{45.6}    & 14.1 & \textbf{30.9}  & 41.8  & \textbf{52.4} & \textbf{1.1}     & 0.6       & \textbf{25.3} & 0.2   & \textbf{20.7}   \\
\hline
\end{tabular}%
}
\label{Table:preclass-sunrgbd-table}
\end{table}

\subsection{Pre-trained Detector}
\label{ssec:pretrained}

We show that jointly training the detector and our proposed augmentor in the pre-training stage can improve the performance.
We compare our method with 3DIoUMatch~\cite{wang20213dioumatch} on both ScanNet~\cite{Dai_2017_CVPR} and SUN RGB-D~\cite{Song_2015_CVPR} with 5\%, 10\%, and 20\% labeled training data.
The results in Table~\ref{Table:pretrain-table} point out that {\ourname} results in larger performance gains when few training data are available, such as 10\% labeled training data on ScanNet and 5\% labeled training data on SUN RGB-D.

\begin{table}[h]
\caption{We report the pre-trained model performance of {\ourname} and 3DIoUMatch with different amounts of labeled training data.
}
\vspace{2mm}
\centering
\scriptsize
\renewcommand{\arraystretch}{1.2}
\resizebox{\columnwidth}{!}{%
\begin{tabular}{cccccccc}
\toprule
\multirow{3}{*}{Dataset} & \multirow{3}{*}{Model} & \multicolumn{2}{c}{5\%} & \multicolumn{2}{c}{10\%} & \multicolumn{2}{c}{20\%} \\
\cline{3-8}
          &            & mAP   & mAP  & mAP   & mAP  & mAP   & mAP  \\
          &            & @0.25 & @0.5 & @0.25 & @0.5 & @0.25 & @0.5 \\
\cline{1-8}
          & 3DIoUMatch~\cite{wang20213dioumatch} & 29.5     & 13.6    & 40.6     & 20.8    & 47.4     & 29.1    \\
ScanNet   & OPA        & \textbf{33.3}     & \textbf{16.2}    & \textbf{45.8}     & \textbf{26.1}    & \textbf{50.2}     & \textbf{31.8}    \\
          & Gain (\%)  & 3.8↑      & 2.6↑     & 5.2↑      & 5.3↑     & 2.8↑      & 2.7↑     \\
\cline{1-8}
          & 3DIoUMatch~\cite{wang20213dioumatch} & 31.0     & 14.5    & 41.5     & 21.4   & 48.0    & \textbf{26.6}      \\
SUN RGB-D & OPA        & \textbf{36.1}     & \textbf{16.3}    & \textbf{44.4}     & \textbf{23.8}   & \textbf{48.5}    & \textbf{26.6}      \\
          & Gain (\%)  & 5.1↑      & 1.8↑     & 2.9↑      & 2.4↑    & 0.5↑     & 0         \\
\bottomrule
\end{tabular}
}
\label{Table:pretrain-table}
\end{table}

\subsection{Displacements Analysis}
\label{ssec:dis}
\begin{figure*}
	\centering
	    \includegraphics[width=0.95\linewidth]{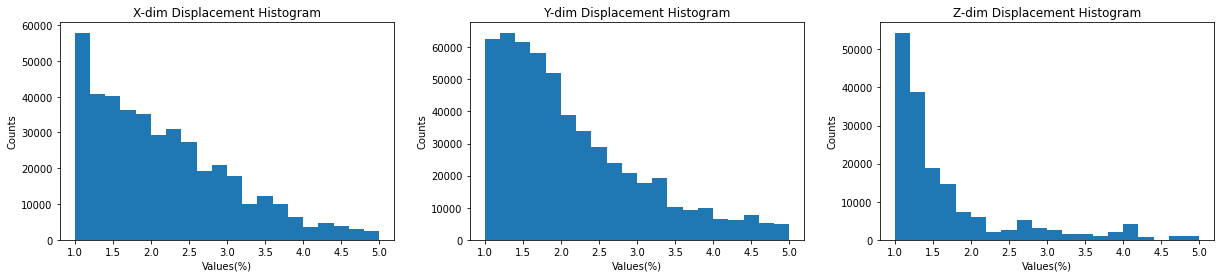}
	\vspace{-3mm}
	\caption{
         Histograms of the displacement distributions of the augmented points along x-direction (left), y-direction (middle), and z-direction (right).
        }
	\label{fig:dis}
\end{figure*}

We analyze how {\ourname} augments the points along x-direction, y-direction, and z-direction.
The three histograms in Figure~\ref{fig:dis} display the displacement distributions along the three directions, respectively.
The x-axis in each histogram represents the ratio of the absolute displacement to its corresponding bounding box dimension while the y-axis gives the frequency.
Here, we only count points with displacement ratios larger than 1\%, which are considered significant augmentation.
This figure shows that our augmentor learns to produce more realistic augmented data.
For example, most points are augmented with displacement ratios less than 3\% along x-direction and y-direction while less than 1.5\% along z-direction. 
It is the expected result since all objects should be placed on the ground or on top of another object due to gravity.
It implies that our augmentor can appropriately augment objects by increasing data variance and maintaining object distinctiveness at the same time.

\subsection{Qualitative Visualization}
\label{ssec:vis}
\begin{figure*}
	\centering
	    \includegraphics[width=0.95\linewidth]{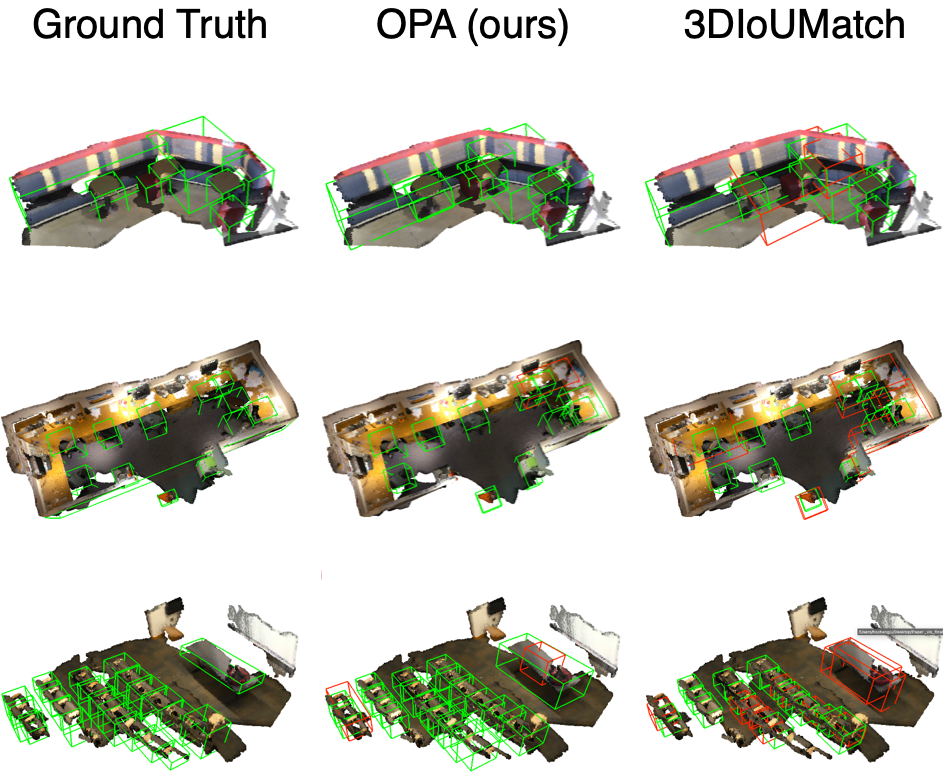}
	\vspace{-3mm}
	\caption{
         Qualitative results on the ScanNet val set, training with 10\% labeled data.
         The green bounding boxes denote the IoU score of proposals greater than 0.25, while the red bounding boxes indicate the IoU score of the proposal less than 0.25. 
        }
	\label{fig:visScanNet}
\end{figure*}

\begin{figure*}
	\centering
	    \includegraphics[width=0.95\linewidth]{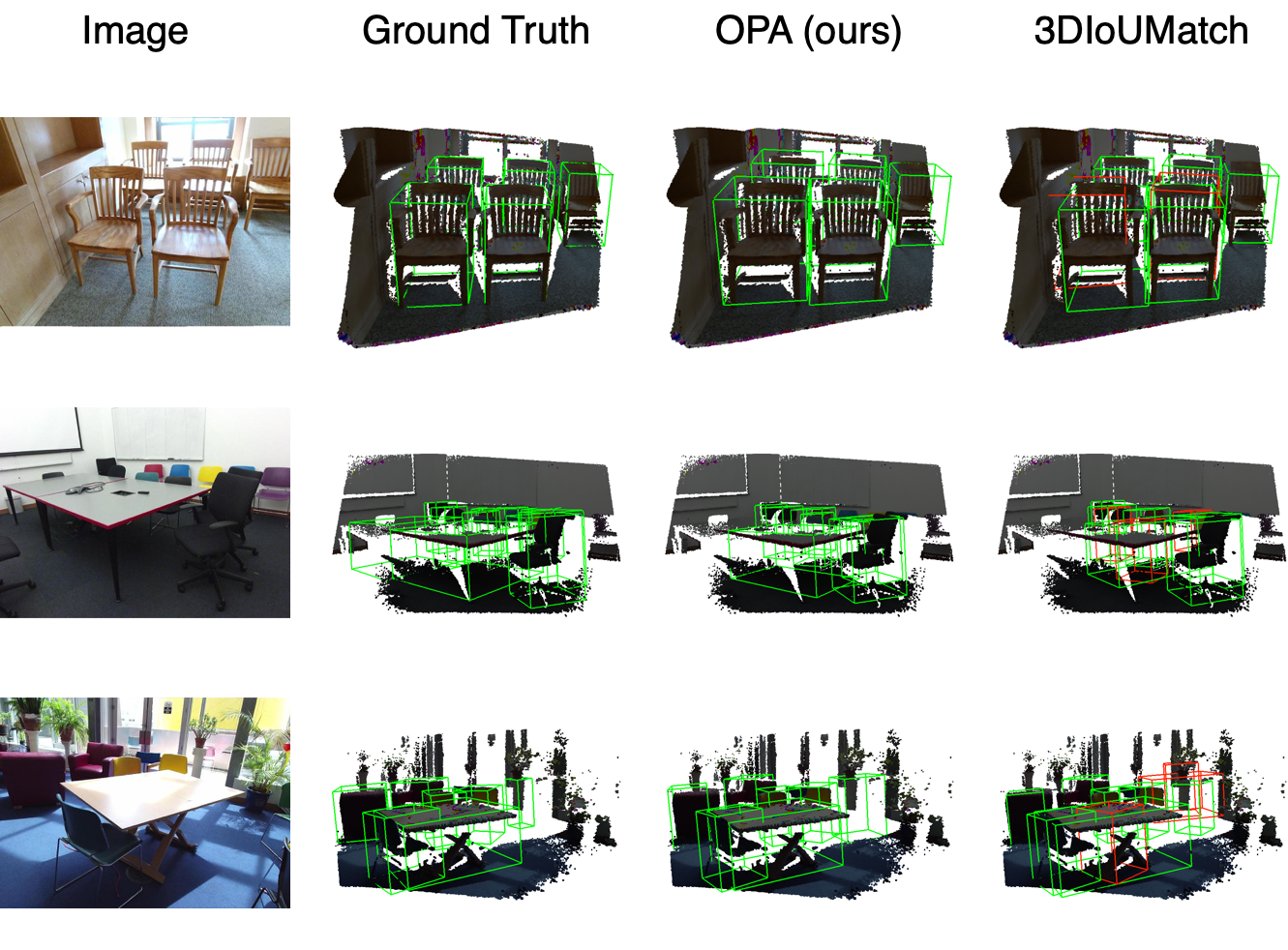}
	\vspace{-3mm}
	\caption{
         Qualitative results on the SUN RGB-D val set, training with 5\% labeled data.
         The green bounding boxes denote the IoU score of proposals greater than 0.25, while the red bounding boxes indicate the IoU score of the proposal less than 0.25.
        }
	\label{fig:visSUNRGBD}
\end{figure*}

We show qualitative results on the validation set of the model using ScanNet~\cite{Dai_2017_CVPR} with 10\% labeled data in Figure~\ref{fig:visScanNet} and on the validation set using SUN RGB-D~\cite{Song_2015_CVPR} with 5\% labeled data in Figure~\ref{fig:visSUNRGBD}.
In the results, the green bounding boxes in the scenes indicate proposals with IoU $\geq$ 0.25, and the red bounding boxes denote proposals with IoU $<$ 0.25.
Overall, our method predicts the objects more precisely compared to ground truths and can locate the objects that are highly occluded by other objects.

\clearpage
\bibliography{egbib}

\begin{thebibliography}{38}
\providecommand{\natexlab}[1]{#1}
\providecommand{\url}[1]{\texttt{#1}}
\expandafter\ifx\csname urlstyle\endcsname\relax
  \providecommand{\doi}[1]{doi: #1}\else
  \providecommand{\doi}{doi: \begingroup \urlstyle{rm}\Url}\fi

\bibitem[Berthelot et~al.(2019)Berthelot, Carlini, Goodfellow, Papernot,
  Oliver, and Raffel]{berthelot2019mixmatch}
David Berthelot, Nicholas Carlini, Ian Goodfellow, Nicolas Papernot, Avital
  Oliver, and Colin~A Raffel.
\newblock Mixmatch: A holistic approach to semi-supervised learning.
\newblock \emph{Advances in Neural Information Processing Systems}, 32, 2019.

\bibitem[Chen et~al.(2020)Chen, Hu, Gavves, Mensink, Mettes, Yang, and
  Snoek]{chen2020pointmixup}
Yunlu Chen, Vincent~Tao Hu, Efstratios Gavves, Thomas Mensink, Pascal Mettes,
  Pengwan Yang, and Cees~GM Snoek.
\newblock Pointmixup: Augmentation for point clouds.
\newblock In \emph{European Conference on Computer Vision}, pages 330--345,
  2020.

\bibitem[Cheng et~al.(2020)Cheng, Leng, Cubuk, Zoph, Bai, Ngiam, Song, Caine,
  Vasudevan, Li, et~al.]{cheng2020improving}
Shuyang Cheng, Zhaoqi Leng, Ekin~Dogus Cubuk, Barret Zoph, Chunyan Bai, Jiquan
  Ngiam, Yang Song, Benjamin Caine, Vijay Vasudevan, Congcong Li, et~al.
\newblock Improving 3d object detection through progressive population based
  augmentation.
\newblock In \emph{European Conference on Computer Vision}, pages 279--294,
  2020.

\bibitem[Choi et~al.(2020)Choi, Song, and Kwak]{choi2020part}
Jaeseok Choi, Yeji Song, and Nojun Kwak.
\newblock Part-aware data augmentation for 3d object detection in point cloud.
\newblock \emph{arXiv preprint arXiv:2007.13373}, 2020.

\bibitem[Choy et~al.(2019)Choy, Gwak, and Savarese]{choy20194d}
Christopher Choy, JunYoung Gwak, and Silvio Savarese.
\newblock 4d spatio-temporal convnets: Minkowski convolutional neural networks.
\newblock In \emph{Proceedings of the IEEE Conference on Computer Vision and
  Pattern Recognition}, pages 3075--3084, 2019.

\bibitem[Dai et~al.(2017)Dai, Chang, Savva, Halber, Funkhouser, and
  Niessner]{Dai_2017_CVPR}
Angela Dai, Angel~X. Chang, Manolis Savva, Maciej Halber, Thomas Funkhouser,
  and Matthias Niessner.
\newblock Scannet: Richly-annotated 3d reconstructions of indoor scenes.
\newblock In \emph{Proceedings of the IEEE Conference on Computer Vision and
  Pattern Recognition}, 2017.

\bibitem[Jeong et~al.(2019)Jeong, Lee, Kim, and Kwak]{NEURIPS2019_d0f4dae8}
Jisoo Jeong, Seungeui Lee, Jeesoo Kim, and Nojun Kwak.
\newblock Consistency-based semi-supervised learning for object detection.
\newblock In \emph{Advances in Neural Information Processing Systems}, 2019.

\bibitem[Kim et~al.(2021)Kim, Lee, Hwang, Lee, Hwang, and Kim]{Kim_2021_ICCV}
Sihyeon Kim, Sanghyeok Lee, Dasol Hwang, Jaewon Lee, Seong~Jae Hwang, and
  Hyunwoo~J. Kim.
\newblock Point cloud augmentation with weighted local transformations.
\newblock In \emph{Proceedings of the IEEE International Conference on Computer
  Vision}, pages 548--557, 2021.

\bibitem[Laine and Aila(2016)]{laine2016temporal}
Samuli Laine and Timo Aila.
\newblock Temporal ensembling for semi-supervised learning.
\newblock \emph{arXiv preprint arXiv:1610.02242}, 2016.

\bibitem[Lee et~al.(2013)]{lee2013pseudo}
Dong-Hyun Lee et~al.
\newblock Pseudo-label: The simple and efficient semi-supervised learning
  method for deep neural networks.
\newblock In \emph{ICML Workshop on Challenges in Representation Learning},
  volume~3, page 896, 2013.

\bibitem[Li et~al.(2021)Li, Wu, Shrivastava, and Davis]{li2021rethinking}
Hengduo Li, Zuxuan Wu, Abhinav Shrivastava, and Larry~S Davis.
\newblock Rethinking pseudo labels for semi-supervised object detection.
\newblock \emph{arXiv preprint arXiv:2106.00168}, 3\penalty0 (5), 2021.

\bibitem[Li et~al.(2020{\natexlab{a}})Li, Li, Heng, and Fu]{li2020pointaugment}
Ruihui Li, Xianzhi Li, Pheng-Ann Heng, and Chi-Wing Fu.
\newblock Pointaugment: an auto-augmentation framework for point cloud
  classification.
\newblock In \emph{Proceedings of the IEEE/CVF Conference on Computer Vision
  and Pattern Recognition}, pages 6378--6387, 2020{\natexlab{a}}.

\bibitem[Li et~al.(2020{\natexlab{b}})Li, Huang, Qin, Wang, and
  Gong]{li2020improving}
Yandong Li, Di~Huang, Danfeng Qin, Liqiang Wang, and Boqing Gong.
\newblock Improving object detection with selective self-supervised
  self-training.
\newblock In \emph{European Conference on Computer Vision}, pages 589--607,
  2020{\natexlab{b}}.

\bibitem[Liu et~al.(2021)Liu, Ma, He, Kuo, Chen, Zhang, Wu, Kira, and
  Vajda]{liu2021unbiased}
Yen-Cheng Liu, Chih-Yao Ma, Zijian He, Chia-Wen Kuo, Kan Chen, Peizhao Zhang,
  Bichen Wu, Zsolt Kira, and Peter Vajda.
\newblock Unbiased teacher for semi-supervised object detection.
\newblock \emph{arXiv preprint arXiv:2102.09480}, 2021.

\bibitem[Qi et~al.(2017{\natexlab{a}})Qi, Su, Mo, and Guibas]{qi2017pointnet}
Charles~R Qi, Hao Su, Kaichun Mo, and Leonidas~J Guibas.
\newblock Pointnet: Deep learning on point sets for 3d classification and
  segmentation.
\newblock In \emph{Proceedings of the IEEE Conference on Computer Vision and
  Pattern Recognition}, pages 652--660, 2017{\natexlab{a}}.

\bibitem[Qi et~al.(2019)Qi, Litany, He, and Guibas]{qi2019deep}
Charles~R Qi, Or~Litany, Kaiming He, and Leonidas~J Guibas.
\newblock Deep hough voting for 3d object detection in point clouds.
\newblock In \emph{proceedings of the IEEE International Conference on Computer
  Vision}, pages 9277--9286, 2019.

\bibitem[Qi et~al.(2017{\natexlab{b}})Qi, Yi, Su, and Guibas]{qi2017pointnet++}
Charles~Ruizhongtai Qi, Li~Yi, Hao Su, and Leonidas~J Guibas.
\newblock Pointnet++: Deep hierarchical feature learning on point sets in a
  metric space.
\newblock \emph{Advances in Neural Information Processing Systems}, 30,
  2017{\natexlab{b}}.

\bibitem[Sajjadi et~al.(2016)Sajjadi, Javanmardi, and
  Tasdizen]{sajjadi2016regularization}
Mehdi Sajjadi, Mehran Javanmardi, and Tolga Tasdizen.
\newblock Regularization with stochastic transformations and perturbations for
  deep semi-supervised learning.
\newblock \emph{Advances in Neural Information Processing Systems}, 29, 2016.

\bibitem[Shi et~al.(2019)Shi, Wang, and Li]{shi2019pointrcnn}
Shaoshuai Shi, Xiaogang Wang, and Hongsheng Li.
\newblock Pointrcnn: 3d object proposal generation and detection from point
  cloud.
\newblock In \emph{Proceedings of the IEEE Conference on Computer Vision and
  Pattern Recognition}, pages 770--779, 2019.

\bibitem[Shi et~al.(2020{\natexlab{a}})Shi, Guo, Jiang, Wang, Shi, Wang, and
  Li]{shi2020pv}
Shaoshuai Shi, Chaoxu Guo, Li~Jiang, Zhe Wang, Jianping Shi, Xiaogang Wang, and
  Hongsheng Li.
\newblock Pv-rcnn: Point-voxel feature set abstraction for 3d object detection.
\newblock In \emph{Proceedings of the IEEE Conference on Computer Vision and
  Pattern Recognition}, pages 10529--10538, 2020{\natexlab{a}}.

\bibitem[Shi et~al.(2020{\natexlab{b}})Shi, Wang, Shi, Wang, and
  Li]{shi2020points}
Shaoshuai Shi, Zhe Wang, Jianping Shi, Xiaogang Wang, and Hongsheng Li.
\newblock From points to parts: 3d object detection from point cloud with
  part-aware and part-aggregation network.
\newblock \emph{IEEE transactions on pattern analysis and machine
  intelligence}, 43\penalty0 (8):\penalty0 2647--2664, 2020{\natexlab{b}}.

\bibitem[Sohn et~al.(2020{\natexlab{a}})Sohn, Berthelot, Carlini, Zhang, Zhang,
  Raffel, Cubuk, Kurakin, and Li]{sohn2020fixmatch}
Kihyuk Sohn, David Berthelot, Nicholas Carlini, Zizhao Zhang, Han Zhang,
  Colin~A Raffel, Ekin~Dogus Cubuk, Alexey Kurakin, and Chun-Liang Li.
\newblock Fixmatch: Simplifying semi-supervised learning with consistency and
  confidence.
\newblock In \emph{Advances in Neural Information Processing Systems},
  2020{\natexlab{a}}.

\bibitem[Sohn et~al.(2020{\natexlab{b}})Sohn, Zhang, Li, Zhang, Lee, and
  Pfister]{sohn2020simple}
Kihyuk Sohn, Zizhao Zhang, Chun-Liang Li, Han Zhang, Chen-Yu Lee, and Tomas
  Pfister.
\newblock A simple semi-supervised learning framework for object detection.
\newblock \emph{arXiv preprint arXiv:2005.04757}, 2020{\natexlab{b}}.

\bibitem[Song et~al.(2015)Song, Lichtenberg, and Xiao]{Song_2015_CVPR}
Shuran Song, Samuel~P. Lichtenberg, and Jianxiong Xiao.
\newblock Sun rgb-d: A rgb-d scene understanding benchmark suite.
\newblock In \emph{Proceedings of the IEEE Conference on Computer Vision and
  Pattern Recognition}, 2015.

\bibitem[Tang et~al.(2021{\natexlab{a}})Tang, Ramaiah, Wang, Xu, and
  Xiong]{tang2021proposal}
Peng Tang, Chetan Ramaiah, Yan Wang, Ran Xu, and Caiming Xiong.
\newblock Proposal learning for semi-supervised object detection.
\newblock In \emph{Proceedings of the IEEE Winter Conference on Applications of
  Computer Vision}, pages 2291--2301, 2021{\natexlab{a}}.

\bibitem[Tang et~al.(2021{\natexlab{b}})Tang, Chen, Luo, and
  Zhang]{tang2021humble}
Yihe Tang, Weifeng Chen, Yijun Luo, and Yuting Zhang.
\newblock Humble teachers teach better students for semi-supervised object
  detection.
\newblock In \emph{Proceedings of the IEEE Conference on Computer Vision and
  Pattern Recognition}, pages 3132--3141, 2021{\natexlab{b}}.

\bibitem[Tarvainen and Valpola(2017)]{tarvainen2017mean}
Antti Tarvainen and Harri Valpola.
\newblock Mean teachers are better role models: Weight-averaged consistency
  targets improve semi-supervised deep learning results.
\newblock In \emph{Advances in Neural Information Processing Systems}, 2017.

\bibitem[Umam et~al.(2022)Umam, Yang, Chuang, Chuang, and Lin]{umam2022point}
Ardian Umam, Cheng-Kun Yang, Yung-Yu Chuang, Jen-Hui Chuang, and Yen-Yu Lin.
\newblock Point mixswap: Attentional point cloud mixing via swapping matched
  structural divisions.
\newblock In \emph{European Conference on Computer Vision}, pages 596--611.
  Springer, 2022.

\bibitem[Wang et~al.(2021)Wang, Cong, Litany, Gao, and
  Guibas]{wang20213dioumatch}
He~Wang, Yezhen Cong, Or~Litany, Yue Gao, and Leonidas~J Guibas.
\newblock 3dioumatch: Leveraging iou prediction for semi-supervised 3d object
  detection.
\newblock In \emph{Proceedings of the IEEE Conference on Computer Vision and
  Pattern Recognition}, pages 14615--14624, 2021.

\bibitem[Wang et~al.(2018)Wang, Yan, Zhang, Zhang, and Lin]{Wang_2018_CVPR}
Keze Wang, Xiaopeng Yan, Dongyu Zhang, Lei Zhang, and Liang Lin.
\newblock Towards human-machine cooperation: Self-supervised sample mining for
  object detection.
\newblock In \emph{Proceedings of the IEEE Conference on Computer Vision and
  Pattern Recognition}, 2018.

\bibitem[Xie et~al.(2020)Xie, Dai, Hovy, Luong, and Le]{xie2020unsupervised}
Qizhe Xie, Zihang Dai, Eduard Hovy, Thang Luong, and Quoc Le.
\newblock Unsupervised data augmentation for consistency training.
\newblock \emph{Advances in Neural Information Processing Systems},
  33:\penalty0 6256--6268, 2020.

\bibitem[Xu et~al.(2021)Xu, Zhang, Hu, Wang, Wang, Wei, Bai, and
  Liu]{xu2021end}
Mengde Xu, Zheng Zhang, Han Hu, Jianfeng Wang, Lijuan Wang, Fangyun Wei, Xiang
  Bai, and Zicheng Liu.
\newblock End-to-end semi-supervised object detection with soft teacher.
\newblock In \emph{Proceedings of the IEEE International Conference on Computer
  Vision}, pages 3060--3069, 2021.

\bibitem[Yan et~al.(2018)Yan, Mao, and Li]{s18103337}
Yan Yan, Yuxing Mao, and Bo~Li.
\newblock Second: Sparsely embedded convolutional detection.
\newblock \emph{Sensors}, 18\penalty0 (10), 2018.
\newblock ISSN 1424-8220.

\bibitem[Yang et~al.(2020)Yang, Sun, Liu, and Jia]{yang20203dssd}
Zetong Yang, Yanan Sun, Shu Liu, and Jiaya Jia.
\newblock 3dssd: Point-based 3d single stage object detector.
\newblock In \emph{Proceedings of the IEEE Conference on Computer Vision and
  Pattern Recognition}, pages 11040--11048, 2020.

\bibitem[Zhang et~al.(2017)Zhang, Cisse, Dauphin, and
  Lopez-Paz]{zhang2017mixup}
Hongyi Zhang, Moustapha Cisse, Yann~N Dauphin, and David Lopez-Paz.
\newblock Mixup: Beyond empirical risk minimization.
\newblock \emph{arXiv preprint arXiv:1710.09412}, 2017.

\bibitem[Zhao et~al.(2020)Zhao, Chua, and Lee]{zhao2020sess}
Na~Zhao, Tat-Seng Chua, and Gim~Hee Lee.
\newblock Sess: Self-ensembling semi-supervised 3d object detection.
\newblock In \emph{Proceedings of the IEEE Conference on Computer Vision and
  Pattern Recognition}, pages 11079--11087, 2020.

\bibitem[Zhou et~al.(2021)Zhou, Yu, Wang, Qian, and Li]{zhou2021instant}
Qiang Zhou, Chaohui Yu, Zhibin Wang, Qi~Qian, and Hao Li.
\newblock Instant-teaching: An end-to-end semi-supervised object detection
  framework.
\newblock In \emph{Proceedings of the IEEE Conference on Computer Vision and
  Pattern Recognition}, pages 4081--4090, 2021.

\bibitem[Zhou and Tuzel(2018)]{zhou2018voxelnet}
Yin Zhou and Oncel Tuzel.
\newblock Voxelnet: End-to-end learning for point cloud based 3d object
  detection.
\newblock In \emph{Proceedings of the IEEE Conference on Computer Vision and
  Pattern Recognition}, pages 4490--4499, 2018.

\end{thebibliography}
\end{document}